# CUDA-Self-Organizing feature map based visual sentiment analysis of bank customer complaints for Analytical CRM

**Rohit Gavval**[1,2]
[2]*School of Computer & Information Sciences*
*University of Hyderabad*
Hyderabad, India - 500046
rohitg6790@gmail.com

**Vadlamani Ravi**[1*]
[1]*Center of Excellence in Analytics,*
Institute for Development and Research in Banking Technology,
Castle Hills Road No. 1, Hyderabad, India - 500057
padmarav@gmail.com

**Kalavala Revanth Harsha1**[1,3]
[3]*Indian Institute of Technology (Indian School of Mines), Dhanbad*
Dhanbad - 826004
kalavalarevanth@gmail.com

**Akhilesh Gangwar**[1]
gangwar.akhilesh1993@gmail.com

**Kumar Ravi**[1]
kumar_ravi66@yahoo.co.in

**Abstract**

With the widespread use of social media, companies now have access to a wealth of customer feedback data which has valuable applications to Customer Relationship Management (CRM). Analyzing customer grievances data, is paramount as their speedy non-redressal would lead to customer churn resulting in lower profitability. In this paper, we propose a descriptive analytics framework using Self-organizing feature map (SOM), for Visual Sentiment Analysis of customer complaints. The network learns the inherent grouping of the complaints automatically which can then be visualized too using various techniques. Analytical Customer Relationship Management (ACRM) executives can draw useful business insights from the maps and take timely remedial action. We also propose a high-performance version of the algorithm CUDASOM (CUDA based Self Organizing feature Map) implemented using NVIDIA parallel computing platform, CUDA, which speeds up the processing of high-dimensional text data and generates fast results. The efficacy of the proposed model has been demonstrated on the customer complaints data regarding the products and services of four leading Indian banks. CUDASOM achieved an average speed up of 44 times. Our approach can expand research into intelligent grievance redressal system to provide rapid solutions to the complaining customers.

*Keywords— Customer Complaints; Analytical CRM; Self-Organizing Map; CUDA; Visual Sentiment Analysis*

* Corresponding Author, Phone: +914023294042; FAX: +9140235351576

1. INTRODUCTION

Organizations embarking on the customer relationship management (CRM) journey are increasingly using customer data to drive business decisions. The Banking, Financial services and Insurance (BFSI) industry also relies heavily on data analytics for their day-to-day functions like fraud detection, credit scoring, risk analytics or cross-sell/up-sell, which come under the realm of Analytical CRM (ACRM). In addition to a large amount of data collected by these organizations from various customer touchpoints, critical customer feedback data is now available from social media platforms, which are the most preferred channels for customers to express their opinions/grievances regarding the products and services.

Efficient analysis of customer complaints can help the organizations strengthen the communication channels with their customers and resolve their issues quickly, turning complaining customers to loyal repeat customers [1], [2]. Essentially, non-redressal of complaints leads to customer churn and decreased sales. Yet, going through the expectedly large amounts of feedback would be time-consuming and would affect the turnaround time. A purely manual approach would be infeasible in most of the cases. Alternative approaches such as text mining and NLP techniques can help greatly here, by providing an automated way of generating insights from the large amounts of text. The most popular task carried out on customer feedback data is the classification of the sentiments expressed by the customer in their respective feedback document. This task involves classifying the sentiments into positive or negative classes to extract the overall perception of the customers towards a product or service. *Sentiment classification* is performed at various levels including document-level, sentence-level, and aspect-level. *Opinion summarization* is another class of analysis of customer feedback, which generates a summary of the feedback for a target (product or service) from a large corpus of feedback documents. Before a full-fledged sentiment analysis of the complaints is undertaken, visualization of the complaints by product and severity category greatly reduces the time and effort for the bank executives.

Therefore, to meet this requirement, we employed self-feature organizing maps, which are unsupervised and non-linear neural networks, in analyzing customer complaints.

Since inception, self-organizing feature maps have been extensively used for a variety of data analysis tasks such as clustering and data visualization in industry, linguistics, finance, and bioinformatics. SOM is a type of artificial neural network that is trained using unsupervised learning to produce low-dimensional representations of the input variables, making visualization of high-dimensional data possible. The standard SOM architecture consists of two layers - the input layer and the output layer. For a dataset with $d$ dimensional vectors, the input layer contains $d$ nodes, each node representing one attribute of a sample. The input layer presents the training data one sample at a time to the output layer, which is a regularly arranged (usually two-dimensional) lattice of neurons that represent the distribution of input data with a smaller set of "prototype vectors". Each output node is fully connected to the input layer and contains a weight vector of the same dimension as that of the input vectors. As the network is trained, samples or records that are more similar get associated with adjacent nodes on the output layer, while less similar ones get associated with distant nodes.

In this paper, we propose an unsupervised learning model, which adequately caters to the need of suitably analyzing customer complaints to derive actionable insights. We analyze four customer complaints datasets related to four leading commercial banks in India using self-organizing maps and demonstrate the efficacy of the proposed method in analyzing and visualizing such data. However, since text data is high dimensional, processing it is computationally expensive and slow. To increase the speed of computation, we also propose CUDASOM, a parallel version of the model which has been implemented on CUDA (Compute Unified Device Architecture) which is a parallel computing architecture developed by NVIDIA.

The paper is structured as follows. First, we present the motivation behind this work in section 2. Then, we highlight the significant contributions of our work in section 3 and present an overview of literature review in section 4. A description of the techniques is provided in section 5. The details of our approach and the experimental setup are explained in section 6 and section 7 respectively. The results are discussed in section 8 and the concluding remarks are presented in section 9.

## 2. MOTIVATION

Sentiment classification of a corpus of complaints would logically assign a negative sentiment score to all the complaints and opinion summarization would result in summaries, which are neither meaningful nor useful. Therefore, while these two tasks are popular in the sentiment analysis domain and are extremely helpful in gaining insights regarding customer perception, they are ineffective in analyzing a corpus of complaints with a view to rapidly draw actionable insights. Moreover, both these techniques are supervised learning techniques which require time and effort for labeling the data and developing a reasonably well-performing model. Although unsupervised learning techniques have been proposed in the literature for sentiment analysis [3][4], these works focus on analysis of datasets containing both positive and negative sentiments whereas the focus of this paper is customer complaints, a type of dataset containing only negative sentiments. On the other end of the spectrum are niche unsupervised techniques like clustering which can deliver quick output, but impose an overhead of painstakingly elaborate analysis in terms of choice of number of clusters and understanding the nature of complaints within a group. Since text data is high-dimensional, visualization of the clusters to learn about possible interrelationships in the data requires help from auxiliary tools. A tool to analyze customer complaints should ideally help the service executives to identify the landscape of complaints so that they could derive actionable insights from them rapidly and also prioritize the complaints for a speedy redressal. The benefits of such tool would be pronounced if the tool provides these insights in a visual format. For example, it would be extremely helpful for a service manager to identify the product-wise distribution of complaints so that the tasks are prioritized accordingly or, to identify a pressing issue affecting a large customer base, so that the team responsible is immediately deployed in resolving the problem. SOM effectively addresses these requirements in the following ways, a) it works in the absence of labels, b) it automatically performs dimensionality reduction making visualization possible c) owing to its topology preservation properties, it helps in visualization of the landscape of the data along with their topological relationships which is paramount and also invaluable for this task. Therefore with SOM at its core, we developed a visual sentiment analysis framework which can help the banks to efficiently handle customer complaints. By the virtue of being an unsupervised approach, the proposed method can quickly paint an approximate picture for the operational CRM staff to start taking action. As such, this method can act as an effective precursor before an in-depth study is performed using the supervised methods.

## 3. CONTRIBUTION

To the best of our knowledge, this is the first effort on employing SOMs for visualizing customer complaints from grievance prioritization and redressal perspective. The major contributions of our work as follows:

- We propose a novel procedure to perform visual sentiment analysis of customer complaints, which could significantly improve the grievance redressal systems. The efficacy of the model has been established with real-world complaints datasets for four Indian banks.
- We also propose CUDASOM, a novel implementation of CUDA accelerated standard SOM. The implementation has been designed to favor high dimensional text data as opposed to earlier proposals which focused majorly on increasing map sizes or number of training examples. It turned out that the proposed CUDA implementation achieved significant speedup as compared to its CPU counterpart.

## 4. LITERATURE SURVEY

Sentiment analysis has been extensively studied by researchers in the past decade and a half, and numerous works have been published in this field. In this study, we are employing self-organizing maps for sentiment analysis. Hence, we surveyed some research related to sentiment analysis performed using self-organizing maps. As this work also proposes a parallel, CUDA-based version of the self-organizing maps, we also review previous efforts in this direction.

Due to their excellent visualization abilities self-organizing maps have been widely adopted in domains like Bioinformatics [5], Geosciences [6], Finance [7] etc. In sentiment analysis research as well, SOMs have been used to study the sentiments of tourists [8], to infer movie characteristics from movie reviews [9], summarize social network conversation regarding wine varieties [10] and in automatic detection of bullying in social media platforms [11]. Reference [12] proposed a two-step approach to study the sentiment of a forum focused on a cancer treatment drug. They manually assigned a sentiment label to each post from the forum and then trained a SOM with a TF-IDF representation of the posts using the SOM Toolbox. The SOM was used to find clusters in the data and to study the correlation of the codebook vectors with the positive and negative opinion. Reference [13] studied the efficacy of SOM for unsupervised and supervised sentiment analysis tasks. They extracted features from movie review dataset in TF-IDF feature representation and selected the top features ranked by the information gain criterion. The dataset was visualized using the Emergent SOM variant of SOM and classifier was trained using the Learning Vector Quantization algorithm. They concluded that the performance of SOM for sentiment analysis tasks is at par with other machine learning algorithms.

Growing Hierarchical Self-organizing Maps (GHSOM) are a class of self-organizing maps which are frequently used for training SOMs with text data. Reference [14] trained a classifier using the Enrich-GHSOM to perform aspect-level classification of product reviews. They modeled the hierarchical relationships between the target entity and the aspects as an ontology and used the hierarchical structure of the Enrich GHSOM to compute the overall sentiment score of the target objects taking into account the sentiment polarity of aspects. The highest learning accuracy reported was 63.15%.

SOM has also been employed to visualize customer feedback to gain insights. Reference [15] proposed to use SOM for visual comparison of customer feedback from a geospatial perspective. They rendered sentiment maps and trained a SOM with the color vectors extracted from each of the maps. The trained map depicted that the sentiment scores and key concepts are associated with the geographic position of the reviewers. Reference [16] proposed a method to visualize customer feedback along with the corresponding evaluation values. They used probabilistic Latent Semantic Indexing to reduce the dimensions of the word frequency vectors and used these features to train a SOM. Each node of a SOM was then assigned a shade based on the evaluation value of that node.

SOM being a type of a neural network is inherently suitable for parallelism. This characteristic has been extensively exploited and a number of parallelization schemes have been proposed for SOM for applications like computer vision and data mining. One of the first proposals of a parallel SOM is seen in [17]. The efficiency of parallel SOM was demonstrated with a MATLAB implementation of the proposed algorithm on a single core CPU. During the same time another software-based parallel implementation of a SOM called parSOM [18] was proposed for high-performance text data analysis. The software was implemented on hardware ranging from a dual processor Celeron system to a 64-processor Cray Origin 2000. Thereafter, parallel implementations were proposed using a variety of hardware and libraries like OpenGL [19], MapReduce-MPI [20], OpenCL [21] and supercomputers and VLSI chips [22].

With the advent of NVIDIA's CUDA architecture, there was an increase General Purpose GPU (GP-GPU) programming as CUDA enables the non-graphic programmers to easily work with GPUs. Reference [23] proposed an implementation based on input data segmentation, and CUDA streams which are sequences of commands that execute in order. They compared the GPU implementation with CPU implementations by varying the SOM network size, input size and the input dimension size. They concluded that speedups due to GPU are not significant in case of small input size or dimension size. The largest dimension tested was 32. Reference [24] evaluated the performance of SOM parallelization on single and multiple GPUs using CUDA, OpenCL and MPI. They reported that on a single GPU, CUDA results in higher speedup as compared to OpenCL, while on multiple GPU setups using CUDA with MPI, they observed that the speedup increases with the number of GPUs for larger datasets but only by a factor of 4. Overall, it was noticed that pure CUDA implementation was the fastest. The largest dimension tested was 8.

Reference [25] proposed a GPU accelerated Batch-SOM for high-dimensional data. The map was organized in a 2D texture format and vertex and fragment shaders were responsible for finding the best matching unit and updating the weights respectively. The authors claimed to achieve a speedup of 15x to 40x in comparison with CPU implementation. The highest dimension tested was 2048.

Reference [26] proposed a parallel implementation of batch SOM by using network and data partitioning methods. The data was partitioned into $M$ maps and each map was broken down to $C$ nodes. Each node was finally broken down to $K$ weight vector dimensions enabling them to parallelize SOM at the level of dimensions and launching $M \times C \times K$ threads. They reported gains up to 20x. However, the largest dimension size tested was 10 and their tests indicated that the gains drop significantly while increasing the dimension size or input size.

A CUDA-based SOM focused on high-dimensional inputs was proposed in [27] A three-step implementation was proposed and the results were reported on synthetic data comprising a maximum of 1000 dimensions.

Recently, [28] proposed a GPU based implementation of sequential SOM. They exploited the cuBLAS library, which contains optimized linear algebra operations for the CUDA architecture. Datasets with vectors of length 16 and 64 were tested.

5. DESCRIPTION OF THE TECHNIQUES USED

*A. Self-organizing Feature Maps*

Inspired by the research on brain maps and neural models of self-organization [29], Kohonen proposed the self-organizing feature maps (SOM) in the 1980s [30]. SOMs are unsupervised artificial neural networks which borrow principles from vector quantization and neurobiology to generate accurate low dimensional representations (feature maps) from high-dimensional data. Most of the real-world data, for example, text data or data from bioinformatics, is high-dimensional and SOMs prove to be excellent tools for dimension reduction and aiding in analysis of such data. SOM achieves this with its property of self-organization which produces a set of prototype vectors such that vectors corresponding to similar points in the input space are topographically closer on the feature map and those corresponding to dissimilar points are farther apart.

The standard SOM architecture consists of two layers – an input layer and an output layer. The input layer consists of $n$ neurons where, $n$ is the number of features in the input dataset, and presents one input vector at a time, to the output layer. While working with text data, the number of neurons in the input layer will be equal to the number of features generated after processing of text data and this layer will present one document at a time to the output layer. The output layer (often) consists of a two-dimensional lattice of neurons. Each of these neurons is mapped to a weight vector of length $n$.

In each iteration of the training procedure, one input vector is presented to the output layer and all the neurons *compete* to get activated. The weight vector which is most similar to the input vector, is determined as the winning neuron. Based on a neighborhood function, the winning neuron then determines a spatial neighborhood, providing a basis for *cooperation* among neighboring neurons. The neurons in the selected neighborhood then *adapt* their weight vectors in proportion to their distance from the winning neuron, with the winning neuron getting the maximum activation. The weight vectors are updated as follows:

$$w_{ij}(t+1) = w_{ij}(t) + h_{ckij}(t)[x(t) - w_{ij}(t)] \qquad (1)$$

Where, $ij$ represent the spatial coordinates of the neuron with which weight vector $w$ is associated, $t$ represents the iteration, $h$ represents a smoothing kernel like Gaussian, $ck$ represent the spatial coordinates of the winning neuron and $x$ represents the input vector.

Using these principles of competition, cooperation and adaption, SOM produces a feature map which accurately represents the input space. For more details on the principles of SOM and tuning it for better results, interested readers are referred to [29].

B. CUDA

Graphics Processing Units (GPUs) are massive arrays of processors which operate in parallel. Although they are computationally weaker than CPUs, due to their low cost, a large of GPU cores can be bunched together to perform parallel operations much faster by increasing the throughput. They were originally designed for graphics based applications like rendering of video game graphics. However, researchers quickly realized the ability to perform independent and parallel operations on the GPU which led to the birth of GPGPU (General-purpose computing on Graphics Processing Units). GPGPU is the use of GPUs for non-graphic purposes by modeling the data as images or other graphical forms. While this approach enabled the use of GPU for other parallel algorithms, modeling the problem in terms of graphics processing was extremely difficult.

To enable users to design their applications without going into the details of graphics processing NVIDIA developed CUDA (Compute Unified Device Architecture) which is an API based on the C language and an architecture for parallel computing on NVIDIA's graphics cards. With CUDA, programmers can develop massively parallel programs for harnessing the power of GPUs without the knowledge of graphics processing terms like vertex shaders and fragment shaders.

The basic flow of a CUDA program is as follows:

1. Load data into the host memory (main memory)

2. Allocate memory for the data in the device memory (global memory of GPU)

3. Copy the data from host memory to device memory

4. Perform computations on the GPU

5. Copy the results to main memory and return the results to the user

The part of an application which runs on the GPU is defined as a special function called kernel. Kernels are run on the GPU by an array of threads on the GPU such that each thread executes the same kernel based on the data assigned to it. Only one kernel can run on a GPU at a time. Threads are arranged in groups called blocks and blocks are grouped into a single grid. Each block of threads have access to fast memory called shared memory and each thread has its own local memory. All CUDA applications can be developed in C/C++. In addition to these languages, some libraries are available in high level languages like Python (PyCUDA) and R (gpuR) as interfaces to the CUDA APIs.

6. **PROPOSED APPROACH**

The proposed approach is divided into four phases viz,. text preprocessing, feature extraction and representation, implementation of CUDASOM, and segmentation of customer complaints using the developed tool. A schematic of the proposed approach is depicted in Fig. 1.

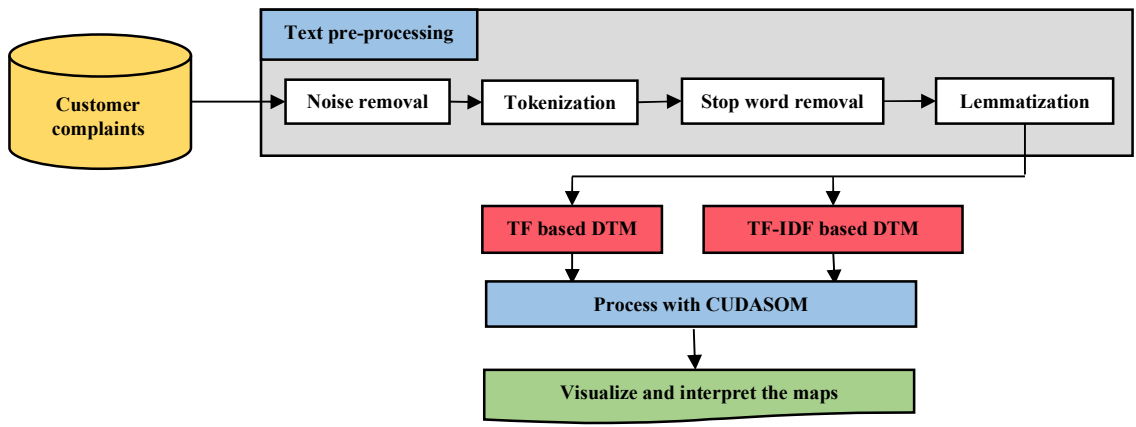

**Figure 1 Schematic of the proposed approach**

*A. Text Preprocessing*

This step was performed using the Natural Language Toolkit (NLTK) in the Python programming language. We followed standard text mining procedures like noise removal, tokenization and lemmatization on the whole corpus. In this study, we rely on the occurrence and significance of terms to capture the severity of complaints as the Euclidean distance metric is used throughout for measuring similarity. Hence, stop words were removed as they contribute to noise in this scenario.

B. *Feature Extraction and Representation*

In this study, we experimented with two feature representations viz. Term Frequency (TF) based DTM and Term Frequency-Inverse Document Frequency (TF-IDF) based Document-Term Matrix (henceforth referred to as DTM).

*1)* *TF based DTM preparation:* This representation involves calculating the term frequencies for the bag-of-words extracted from the text corpus.

*2)* *TF-IDF based DTM preparation:* This representation is a normalized version of the TF based DTM. Each of the term frequencies calculated in TF based DTM is weighted by the inverse of the occurrence of the term in the entire corpus (IDF). Formally, IDF value of a term *t*, appearing in a document *D*, is calculated using equation (2).

$$IDF(t, D) = \log \frac{N}{|\{d \in D : t \in d\}|} \qquad (2)$$

C. *Implementation of CUDASOM*

Several CUDA based SOM implementations have been proposed in the literature. However, most of the recent proposals have focused on the batch version of SOM. The GPU based implementation of standard SOM in [28] shows promising performance. In their proposal, the second dimension of block size is equal to the dimension of input vectors. Although this leads to dimension level parallelism and can contribute to significant gains, this design inherently cannot support high-dimensional input as it is limited by the maximum number of threads supported by a block as per the architecture on which the program is being implemented.

We propose a novel scheme to parallelize the standard SOM on CUDA, with a focus on high dimensional text data. In this scheme, the block size is independent of the length of the data vectors making it suitable for working with DTMs. Our implementation assumes a two dimensional map and hence a three dimensional data structure is used to represent the map along with its weight vectors. Initially, we transform the multi-dimensional data and map structures into single dimensional arrays to improve memory accesses on the GPU. All the data transformations are performed within the memory associated with the GPU (henceforth referred to as the device memory) and then the results are ported back to the main memory. This approach significantly eliminates delays due to host-device memory transfers and synchronizations. We also make use of shared memory on the GPU, in addition to global memory, wherever constant data is repeatedly used for comparisons. The training is performed by three kernels.

*1)* *getDistances kernel:* In each iteration, an input sample is randomly selected on the host and the corresponding index is passed to this kernel call. This kernel distributes a subset of map units in each block and the threads in each block represent the map units. The input vector is copied to the shared memory for each block to minimize global memory access. Within each block, the distance between the input vector and the map units represented by the threads in that block is computed and stored in the shared memory. Using a parallel reduction within the kernel, the minimum distance and the index of the corresponding map unit are stored on the global memory in an array named *bmu*. Hence, each of the blocks stores a possible candidate for best matching unit for the corresponding input.

*2)* *reduceMin kernel*: This kernel uses a parallel reduction scheme on *bmu* to compute the best matching unit. To ensure proper reduction, this kernel is launched with a number of blocks equal to the power of 2.

*3)* *updateWeights kernel:* This kernel computes the distances between the map units using their corresponding matrix coordinates. As hexagonal maps are more accurate and better suited for visualization, we implemented CUDASOM with hexagonal neighbors and accordingly hexagonal coordinates were calculate before computing the distances. The updated weights are written directly to the weights array in the device memory without copying them to the main memory. The weights are copied to the main memory only after all the iterations are completed.

*D. Segmentation of customer complaints using SOM*

In this phase, we trained CUDASOM with each of the dataset separately. The calculation of the number of iterations and the map size for each dataset, was inspired by the heuristics adopted in the SOM Toolbox [31]. Fig. 2 presents an outline of the procedure for calculating the number of iterations and the map size.

Based on the guidelines in [29], we initialized the prototype vectors using the regular, two-dimensional sequence of vectors taken along a hyperplane spanned by the two largest principal components of the DTM. To select the initial learning rate parameter, we performed a series of experiments on benchmark datasets [32] like Iris and Breast Cancer Wisconsin, and fixed the value at 0.1. The Gaussian decay was used to smooth the learning rate and neighborhood radius.

It has been strongly established in information retrieval studies that cosine similarity, which employs dot product, is a more effective similarity measure for analyzing and clustering text documents [33]. According to [29], if the input is normalized to constant length, and if the vector is high dimensional (which is true in our case), the difference in SOMs based on Euclidean distance and dot product is insignificant. As we employed Euclidean distance in our implementation, we normalized the input data before presenting to SOM.

**Begin**

1. *dataset = D, number of records in the dataset = m, number of dimensions = n, number of neurons in the output layer = munits, number of rows in output layer = nrows, number of columns in output layer = ncols, number of iterations = numItr*

2. Calculate the number of neurons in the output layer as $munits = 5 \times \sqrt{m}$

3. Calculate the principal components from D and sort them in descending order of their corresponding eigen values

4. Extract the first two principal components, $pc_1$ and $pc_2$

5. if $pc_1 = 0 \parallel pc_2 \times munits < pc_1$

6.     $r = 1$

7. else

8.     $r = \sqrt{\dfrac{pc_1}{pc_2}}$

9. $size_1 = \min\left(munits, \sqrt{\dfrac{munits}{r \times \sqrt[3]{\frac{3}{4}}}}\right)$

10. $size_2 = \left\lceil \dfrac{munits}{size_1} \right\rceil$

11. $nrows = \min(size_1, size_2)$

12. $ncols = \max(size_1, size_2)$

13. $nn = nrows \times ncols$

14. $mpd = \dfrac{nn}{m}$

15. $numItr = \lceil 50 \times mpd \rceil \times m \times 4$

16. Return $nrows, ncols, numItr$

**End**

**Figure 2 Calculation procedure of map dimension and number of iterations**

An outline of the procedure for calculating the map dimensions and number of iterations is depicted in Figure 2. We trained CUDASOM with the TF and TF-IDF based DTMs separately for each dataset. The trained prototype vectors were then visualized with a similarity coloring scheme [34]. For each experiment, the quantization error and speed were computed. To ascertain that CUDASOM is producing the correct results, we performed experiments on all datasets using the same settings with a CPU implementation of the SOM.

## 7. EXPERIMENTAL SETUP

We demonstrated the effectiveness of the proposed CUDASOM on customer complaint datasets of four leading Indian banks. The customer complaints were posted on an online public complaint forum. Details of the datasets and the experimental environment are presented as follows.

*A. Dataset details*

The datasets were obtained from [35]. The authors crawled the web and collected 749, 1108, 758, and 1123 consumer complaints and bank executive responses from www.complaintboard.in on four banks namely Axis Bank, HDFC Bank, ICICI Bank, and SBI respectively. The first three are leading Indian private sector banks, whereas SBI is the biggest public sector bank in assets. The datasets were prepared during the period of October-December 2014 and hence they contain complaints posted before December 2014, on the specified forum. Details of the datasets are presented in Table 1.

**Table 1 Details of Datasets**

| Bank Name | No. of documents | No. of complaints considered | Features Generated |
|---|---|---|---|
| Axis Bank | 749 | 513 | 3917 |
| HDFC Bank | 1108 | 637 | 4545 |
| ICICI Bank | 758 | 440 | 4268 |
| SBI Bank | 1123 | 676 | 4470 |

B. *Preprocessing steps*

As the data obtained was raw data, we performed the required preprocessing to prepare the data for analysis. Firstly, we dropped bank executive responses, uninformative documents, duplicates and customer requests to make sure that the dataset uniformly represents customer complaints. Secondly, the data was cleansed to remove noisy features like contact information, email metadata, special characters and words of length less than 3. We also corrected the misspelled words. After the preprocessing, 513, 637, 440, 676 complaints were considered for Axis, HDFC, ICICI and SBI banks respectively.

C. *CUDA setup*

All the experiments have been performed on a computer equipped with Intel® Xeon (R) E5-2640 v4 CPU clocked at
2.4 GHz, 32 GB RAM and an NVIDIA® Quadro® P5000 GPU with 16 GB RAM. The GPU is powered by the NVIDIA Pascal architecture and is equipped with 20 microprocessors holding 128 CUDA cores each. We used CUDA v8.0 with nvidia-384 drivers installed on Ubuntu v16.04 operating environment.

All the models were implemented in Python v3.5.2 and the GPU programming was implemented using the PyCUDA [36] module which helps in accessing NVIDIA's CUDA parallel computation API from Python.

Several implementations [37][38] of SOM employing industrial-grade machine learning frameworks like Tensorflow are available on the open source platform. These software provide vanilla implementations of SOM and do not employ the heuristics proposed in the literature for proper training of SOM. Moreover, they do not provide visualization capabilities like similarity coloring which are more intuitive for visualizing a dataset of customer complaints. Also, as machine learning frameworks are positioned on top of several layers of abstraction they leave very little scope for tuning the core architecture of SOM, As such these implementations were not readily suitable for this study

Hence, we selected PyCUDA as the base framework and implemented the architecture from scratch, incorporating the recommended heuristics and visualization tools suitable for this application.

## 8. RESULTS AND DISCUSSION

As this work proposes two ideas, we discuss the results of each of the ideas separately.

### A. Segmentation of customer complaints using CUDASOM

We evaluated the results based on two soft criteria – the ability of the map to provide a visual aid in the analysis of the complaints and effective topology preservation of the data by CUDASOM, i.e., complaints related to one, or similar products get mapped to neighboring nodes.

Firstly, we compared the results across the two feature representations adopted. Fig. 3 depicts the map generated by the TF based DTM of HDFC dataset and the map generated by the TF-IDF based DTM of the same dataset.

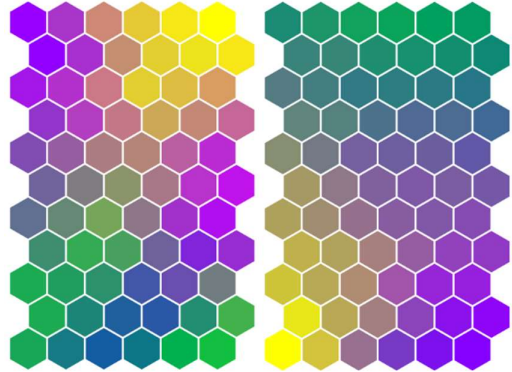

**Figure 3 Maps obtained using TF and TF-IDF representations of HDFC bank dataset respectively**

- The map obtained by the TF-IDF feature representation depicts clearly distinct regions which are uniformly arranged, while the map obtained by the TF representation depicts split, non-uniform regions, which makes interpretation difficult.
- The complaints mapped to the nodes of the map obtained using the TF-IDF representation also demonstrated strong topology preservation. All the complaints related to *loans* are mapped to the nodes in purple (bottom right region of the map) and within them complaints related to *car loans* are mapped to the same or neighbouring nodes. This property could not be observed in the map obtained by the TF representation. On the map, the regions in purple represent different complaints in different regions of the maps and similar complaints are not always mapped to the same region or same node. These two observations are true even for the other three datasets. Therefore, TF-IDF emerged as a better representation technique. It is because of the richness of the information considered by the TF-IDF vis-à-vis TF.

Secondly, we performed an in-depth analysis of the feature maps obtained using the TF-IDF based DTMs of all the four datasets. We analyzed the results to determine if CUDASOM is able to a) produce a product-wise segmentation of the complaints and b) produce a severity-wise segmentation of the complaints. In order to do this, we assigned each document vector to the best matching vector on the trained map. Then, the product-wise segmentation was verified by retrieving the documents mapped to the nodes on the map and studying the distribution of the complaints. In order to verify whether the algorithm can group the complaints based on severity we proceeded as follows. Complaints were labelled as "moderate" or "severe" (symbolically as 1 or 2 respectively) independently by the first author along with two additional human annotators and final labels were assigned by performing a simple majority voting of three sets of labels. After training CUDASOM with unlabeled data and obtaining the feature map, we programmatically marked each neuron of the feature map with the labels of the documents, provided by the human annotators, mapped to it. Broadly, it was noticed that CUDASOM was able to achieve product-wise arrangement of the complaints along with the corresponding severity tagged thereof. The following observations were drawn from the maps:

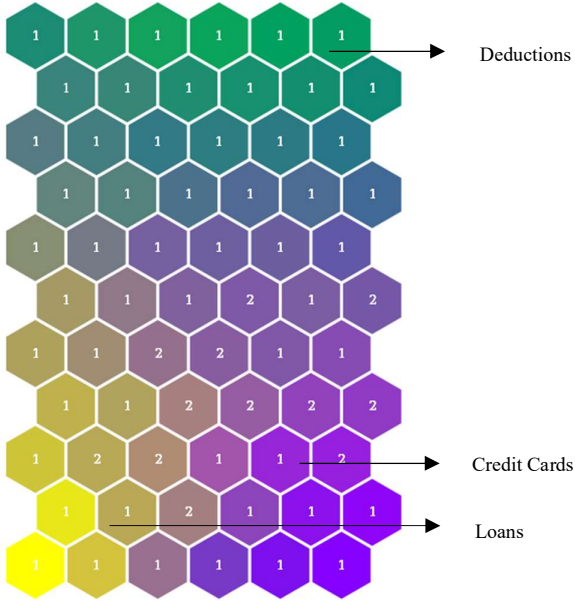

**Figure 4 Map obtained using the TF-IDF representation of HDFC bank dataset**

- The map corresponding to HDFC bank dataset (seen in Fig. 4) depicts three different regions – green, yellow and violet. The green region represents all the complaints related to unexplained deductions in bank accounts which are usually charged by the bank on failure to maintain quarterly average balance or, as annual maintenance balance and is not noted by the customer. The yellow region corresponds to credit cards and violet region corresponds to complaints related to loans. Within the nodes representing loans, moderate complaints about NOC not being sent out on repayment of loans or incorrect information and severe complaints from frustrated customers who are waiting for the loan to be disbursed are well segregated, demonstrating CUDASOM's ability in further segmenting complaints related to a product based on the severity of complaints. Within a region, different shades of the same color represent different aspects of the product complaints. The two bright yellow nodes on the bottom right region of the map correspond to complaints regarding *credit card disapproval*. Some complaints which pertain to less frequent products like *customer care* are in the intersection of the three regions.

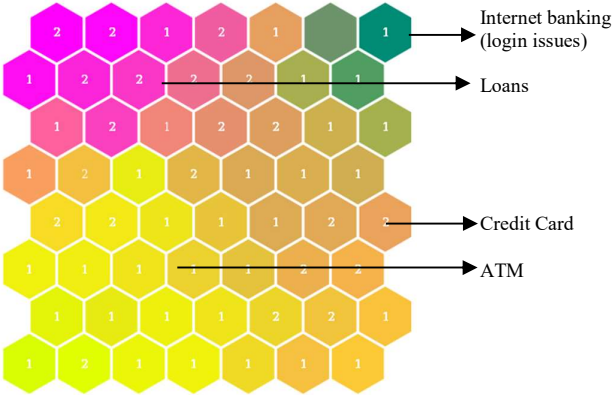

**Figure 5 Map obtained using the TF-IDF representation of Axis bank dataset**

- The map corresponding to Axis bank dataset (seen in Fig. 5) depicts four different regions. The top left region in pink represents *loan* related complaints and the green region on the right represents a strikingly homogenous set of complaints all related to a *login issue* on the bank's website. The region in orange below it represents *credit card* related complaints. As depicted in the case of HDFC dataset, we can observe distinct nodes linked to severe and moderate complaints within the region representing a single product – *credit card*. Severe complaints like incorrect billing and about recovery calls and moderate complaints like different card being sent out have been distinguished by CUDASOM. The region in the bottom left section depicts complaints related to *ATM*.

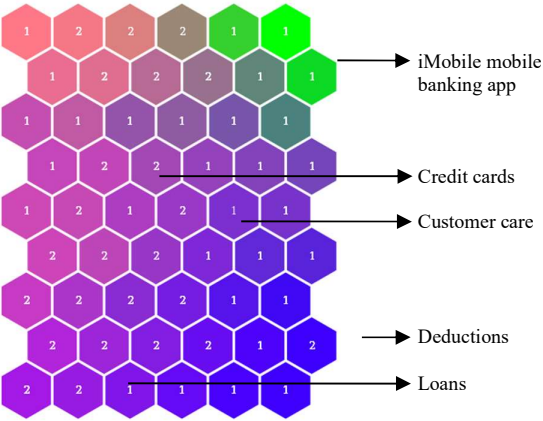

**Figure 6 Map obtained using the TF-IDF representation of ICICI bank dataset**

- The map corresponding to ICICI bank dataset (seen in Fig. 6) depicts roughly four to five different regions in which three regions are distinct but the remaining regions are ambiguous and cannot be distinguished visually. Complaints regarding failure of recharge from the bank's *mobile banking app* are mapped to the green region in the top right section of the map. The bottom right region in blue corresponds to charges and deductions from *bank accounts*. In this region, moderate complaints or enquiries about uninformed deductions and severe complaints threatening to leave the bank due to the same reason have been linked to separate nodes on the map. The region in light pink and dark pink corresponds to complaints related to *credit card* and specifically regarding calls from *recovery agents* or *fraud calls*. The region in the bottom left corresponds to *loan* and the central portion of the map corresponds to complaints regarding *customer care*. It is important to note that although the left half of the map along with the central portion is not visually distinguishable into separate products, even in this map, the complaints have been accurately grouped by SOM in a topology preserving manner. The regions could be indistinguishable in color due to similar wording in the sentences.

- The map corresponding to SBI bank dataset (seen in Fig. 7) depicts four different regions. The top left region in yellow represents *mobile and internet banking*. SOM also remarkably segregated sub-clusters within this region. The bright yellow nodes in the top left corner represent a series of complaints about failed *prepaid phone recharges* made from the *mobile banking app*. The top right section in pink represents complaints about *ATM* transactions. The purple region below it mostly represents complaints about *loans* and few infrequent complaint classes like complaints about deposit of pension in the *accounts*, service at the *branch* etc. The bottom section in blue completely represents *credit card* related complaints. CUDASOM clearly separates severe complaints from moderate complaints in this region. Moderate complaints like customers not being able to generate a PIN for their credit card or not receiving their card on time are mapped to the nodes on the left. Within this section, the region on the bottom left corner in green is related to *Credit card Protection Plan* (CPP), a credit and debit card protection service and mostly moderate complaints are associated with this product. On the other hand, more severe complaints like instances of harassment by recovery team, credit card fraud, etc., are mapped to the nodes on the right, which are labeled as severe.

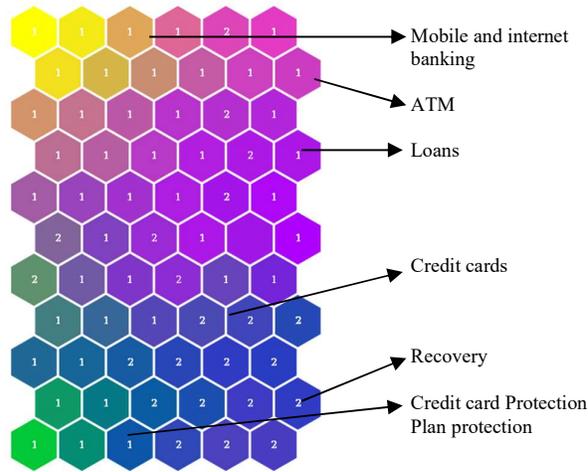

**Figure 7 Map obtained using the TF-IDF representation of SBI dataset**

Descriptive analytics of customer complaints by the proposed model, therefore, provides the advantages of identifying the different products/services which the customer is complaining of, the impact of the affected product/service, any trending issues suddenly affecting a large customer base. This will help the bank executives in charge of customer relationship management department in prioritizing the redressal of complaints.

B. *Performance of CUDASOM*

As mentioned in section 6, to confirm that the results of CUDASOM are accurate, they were compared with the results of experiments performed using the CPU implementation of SOM on all datasets using the same settings. We verified the accuracy by comparing the quantization error and the map generated in the respective experiments. Quantization error is used as a metric to tune the map and as such can be compared only when the dataset and the map sizes are the same.

Table 2 Quantization Errors and Execution Times

|  | **Dataset** | **Dataset dimensions** | **Quantization Error (GPU)** | **Quantization Error(CPU)** | **Time (GPU) in secs** | **Time (CPU) in secs** | **Speed up** |
|---|---|---|---|---|---|---|---|
| **Term Frequency based DTM** | Axis | 513x3917 | 0.80627 | 0.80996 | 22.12 | 915.08 | 41.37 |
|  | HDFC | 637x4545 | 0.80653 | 0.81383 | 32.57 | 1380.82 | 42.40 |
|  | ICICI | 440x4268 | 0.81625 | 0.82011 | 19.94 | 837.02 | 41.98 |
|  | SBI | 676x4470 | 0.82042 | 0.82578 | 27.25 | 1194.57 | 43.84 |
| **TF-IDF based DTM** | Axis | 513x3917 | 0.91366 | 0.91651 | 22.63 | 960.74 | 42.45 |
|  | HDFC | 637x4545 | 0.91562 | 0.92056 | 31.48 | 1412.07 | 44.86 |
|  | ICICI | 440x4268 | 0.91467 | 0.92026 | 23.66 | 985.47 | 41.65 |
|  | SBI | 676x4470 | 0.91769 | 0.92182 | 26.73 | 1254.52 | 46.93 |

Table 3 Execution Time Ratios with Increasing Map Sizes

|  |  | Map Size | | | | | |
|---|---|---|---|---|---|---|---|
|  |  | *16x16* | *32x32* | *64x64* | *128x128* | *256x256* | *512x512* |
| **Reference [28]** | Time (in secs) | 15.50 | 31.20 | 92.30 | 331.00 | 1270.00 | 4900.00 |
|  | Ratio of increase in time | - | 2.01 | 2.96 | 3.59 | 3.84 | 3.86 |
| **CUDASOM** | Time (in secs) | 34.25 | 32.81 | 33.37 | 38.40 | 111.03 | 431.38 |
|  | Ratio of increase in time | - | 0.96 | 1.02 | 1.15 | 2.89 | 3.89 |

● As seen in Table 2, the difference in the quantization errors between CPU and GPU implementations is very small. Differences of this magnitude are expected even in multiple runs of the same program as the initializations and the order in which the samples are presented to SOM would be different. The map obtained from CUDASOM (seen in Fig. 5) and the map obtained from the serial version (seen in Fig. 8) also differs very slightly due to the same reason. An average speed up of 43x is obtained with CUDASOM as compared to the CPU implementation of SOM. The minimum speedup achieved was 41x and the maximum speedup was about ~47x. Although it is a standard practice to compare the time taken by the GPU application with the CPU version, we believe it is unfair as the optimal implementation of CPU is subjective and debatable [27]. As compared to [27] our algorithm has been able to achieve higher peak speedup on input data of 3-fold higher dimensions and the results have been validated on a real-world dataset. To establish that our algorithm performs well with larger map sizes, we ran tests similar to [28] and compared the ratio of time taken with an increase in dimension size of the map. Same data size and number of epochs have been considered. The length of the weight vectors considered is 64. It can be noticed from the table that the ratio of time taken with an exponential increase in map size increases gradually in case of our implementation as compared to [28].

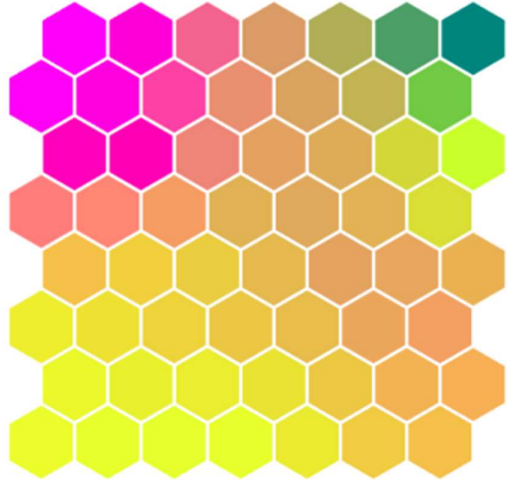

**Figure 8 Map obtained using the TF-IDF representation of Axis bank dataset with standard SOM**

● One limitation of the proposed algorithm is that it is limited by the size of the device memory. Before the training starts, the initialized weight vectors of the map and the data matrix is copied to the device memory and is copied back to the main memory only after the execution of the training phase is completed. Although this approach drastically reduces the CPU-GPU data transfer overhead and synchronization, it limits the size of the data that is processed. In future we will work on alleviating this limitation.

## 9. CONCLUSIONS AND FUTURE DIRECTIONS

In this study, we propose a visual sentiment analysis framework by employing self-organizing maps for analysis of customer complaints. We compared the performance of the model with two vector space model representations of the documents, viz., term frequency representation and TF-IDF representation. We also propose a parallel version of the algorithm for faster processing of complaints and compared its performance with the serial version. The model generated a more distinct segmentation with TF-IDF based DTM as compared to the term frequency based DTM. Visualization depicts that CUDASOM is effectively mapping the complaints related to different products/services into distinct groups. The algorithm also demonstrates a significant speedup as compared to the serial version and can be clearly considered as useful for faster analysis of data with large dimensions.

In future, we will study the performance of the model with more advanced feature representations like Paragraph Embedding and hashing TF-IDF. We are also interested in studying the performance of the model with larger datasets.

ACKNOWLEDGMENT

We thank Mr. Harshal Jaiswal and Mr. Pulipaka Sai Ravi Teja for sparing their valuable time to help us with the annotation task.